\documentclass[journal]{IEEEtran}

\usepackage{graphics} 
\usepackage{blindtext,graphicx,booktabs,amsmath,amssymb,tabularx,tabu,mathrsfs}
\usepackage[utf8]{inputenc}
\usepackage[english]{babel}
\usepackage{url}
\usepackage{color, colortbl}
\usepackage{balance}
\usepackage{xcolor}
\usepackage{algorithm}
\usepackage[noend]{algpseudocode}
\usepackage{multirow}
\usepackage{bbding}
\usepackage{subfig}
\usepackage{graphicx}
\usepackage[noend]{algpseudocode}
\usepackage{subfig}

\ifCLASSINFOpdf
\else
\fi

\begin{document}

\title{Dynamically Local-Enhancement Planner for Large-Scale Autonomous Driving}

\author{Nanshan Deng, Weitao Zhou, Bo Zhang, Junze Wen, Kun Jiang, Zhong Cao, Diange Yang
\thanks{Nanshan Deng, Weitao Zhou, Junze Wen, Kun Jiang, Diange Yang are with the School of Vehicle and Mobility, Tsinghua University, Beijing, China.

Bo Zhang is with Didi Global, China.

Zhong Cao is with Department of Civil and Environmental Engineering, University of Michigan.

W. Zhou and D. Yang are the corresponding authors. (zhouwt, ydg@mail.tsinghua.edu.cn)}

}

\markboth{IEEE ROBOTICS AND AUTOMATION LETTERS}%
{Shell \MakeLowercase{\textit{et al.}}: Bare Demo of IEEEtran.cls for IEEE Journals}

\maketitle

\begin{abstract}

Current autonomous vehicles operate primarily within limited regions, but there is increasing demand for broader applications. However, as models scale, their limited capacity becomes a significant challenge for adapting to novel scenarios. It is increasingly difficult to improve models for new situations using a single monolithic model. To address this issue, we introduce the concept of dynamically enhancing a basic driving planner with local driving data,  without permanently modifying the planner itself. This approach, termed the Dynamically Local-Enhancement (DLE) Planner, aims to improve the scalability of autonomous driving systems without significantly expanding the planner's size. Our approach introduces a position-varying Markov Decision Process formulation coupled with a graph neural network that extracts region-specific driving features from local observation data. The learned features describe the local behavior of the surrounding objects, which is then leveraged to enhance a basic reinforcement learning-based policy. We evaluated our approach in multiple scenarios and compared it with a one-for-all driving model. The results show that our method outperforms the baseline policy in both safety (collision rate) and average reward, while maintaining a lighter scale. This approach has the potential to benefit large-scale autonomous vehicles without the need for largely expanding on-device driving models.

\end{abstract}


\IEEEpeerreviewmaketitle

\section{Introduction and Motivation}

Autonomous driving systems has achieved remarkable progress in recent years \cite{xu2021system},
with companies like Waymo and Cruise demonstrating the ability to operate over 10,000 miles without disengagement \cite{DisReport}. 

Although current road tests are conducted primarily in specific regions or road types, the ultimate goal is to enable large-scale deployment. However, significant variations in driving features across regions pose considerable challenges. The performance of autonomous vehicles may degrade when operating in regions with different driving characteristics. Waymo's safety report underscores the considerable effort required to familiarize autonomous systems with new regulations, road rules, and local driving styles before entering new areas. Similarly, Tesla claims that its FSD autonomous driving system encountered challenges with traffic rules when adapting to Chinese roads.

As the driving region expands, the failure rate increases, and maintaining consistent, non-conflicting decision-making logic becomes progressively more challenging. This issue raises concerns about the adaptability of autonomous systems and impedes their broader application. The goal of this work is to address the adaptability challenges of autonomous vehicles in large-scale environments.

One intuitive approach is to continually enhance the capacity of autonomous driving models to cover more driving regions. However, large-scale applications impose significantly higher requirements on these models. For instance, as the training dataset expands, the model size tends to increase substantially \cite{ouyang2022training}. A similar issue arises when relying on a single model to handle all driving regions. In rule-based systems, this challenge manifests as contradictions among existing rules. Introducing a new rule for a novel scenario requires ensuring there is no conflict with previously established rules, a task that becomes increasingly difficult as the number of rules grows.

There are two primary solutions to this problem: increasing model capacity or restructuring scenarios for simplification. Model enlargement methods include meta-learning, transfer learning, and exploring novel neural network structures, which will be briefly discussed in Section II. However, these methods still fall short of delivering satisfactory real-world performance.

In practice, many companies focus on scenario restructuring, employing strategies such as simplified road structures \cite{Darpa2008}, categorized scenarios, and abstract driving behaviors. For example, planners use Frenet coordinates to represent lanes, treating all lanes as straight with uniform width. Some planners adopt scenario classification approaches, creating dedicated planners for specific driving conditions, such as highway or intersection driving \cite{2020Learning}. Additionally, approaches based on finite state machines (FSM)  define abstract behaviors to adapt to new driving scenarios. However, these simplifications still rely on the goal of using a single model for all driving scenarios and may introduce safety risks by overlooking regional differences \cite{ zhou2022long}.

Our approach stems from the observation that autonomous driving systems exhibit strong regional differences. For example, a vehicle operating in China does not need to incorporate driving data from the U.S. Similarly, a driver relocating to a new city only requires minor adjustments to their driving style for adaptation.

Motivated by this insight, we propose a dynamic local-enhancement planner (DLE) that enhances a basic planner with local driving data. This approach adapts to regional variations without significantly increasing the onboard computational burden. The key contributions of this work are as follows:

\begin{itemize}

\item We propose a Dynamic Local-Enhancement Planner (DLE) framework that adapts autonomous vehicles to regional driving styles by augmenting a base policy with local driving data, eliminating the need for significant scaling of model capacity.

\item We propose a graph-based local feature extraction method that dynamically captures local driving patterns through hierarchical spatiotemporal representation, integrating road structure for local traffic behavior modeling and storage.

\item We develop a reinforcement learning-based policy enhancement method that dynamically optimizes basic driving policy through regional feature integration, enabling autonomous vehicles to autonomously adjust decision-making mechanisms according to localized environmental characteristics, thereby achieving cross-regional behavioral adaptability and performance improvement.
\end{itemize}

\section{Related Works\label{sec:1}} 
Due to the lack of stability in driving policies, particularly in regions with dynamic changes, researchers often rely on approximation, transformation, or simplification methods. These approaches reframe the problem into more manageable forms, enabling the exploration of effective decision-making strategies in dynamic driving environments while alleviating the computational challenges posed by the original problem.

This section reviews existing methods aimed at scaling autonomous driving by enlarging driving models. These methods can be broadly categorized into three types: improving robustness, transfer learning, and representation learning.

\subsection{Improving robustness}

Improving the adaptability of autonomous driving algorithms is crucial for ensuring robust decision-making in diverse environments. Methods like the Reachable Set \cite{althoff2009safety} and the Responsibility Model \cite{zhou2022dynamically} enhance safety by adopting conservative driving policies focused on physical constraints. However, this increased conservatism can limit the set of feasible actions, reducing flexibility. Other approaches improve adaptability by introducing noise during training or by incorporating environmental assumptions. A common technique for simplifying planning is unifying the planner’s coordinate system, such as using the Frenet coordinate system. While this simplifies planning, human driving behavior varies significantly across regions \cite{angkititrakul2012impact}, and generalized policies based on such assumptions can conflict with human cognition, leading to poor performance in unfamiliar traffic environments.

\subsection{Transfer Learning }

Recent studies have approached autonomous driving as a collection of distinct Markov Decision Process (MDP) problems, each tailored to specific regions with different state and action spaces. Transfer learning in reinforcement learning (RL) aims to improve decision-making efficiency by transferring knowledge across tasks. This approach addresses the challenge of maintaining stability in RL performance across various driving scenarios, including urban \cite{2019Model}, highway \cite{2018Formulation}, and other settings.

To mitigate sparse reward problems, reward shaping \cite{harutyunyan2015expressing} is employed, where external knowledge is incorporated as a latent function to augment rewards. Policy distillation \cite{teh2017distral} is also used to minimize the distributional difference between teacher and student policies. Meta-reinforcement learning \cite{finn2017model}, which uses a meta-network to optimize parameters, enables agents to quickly adapt to different tasks. Enhancements to training data, such as adding expert demonstrations \cite{2016End} or adjusting data distribution based on expert knowledge \cite{2014Boosted}, have also been shown to improve decision-making performance. This paper focuses on leveraging regional data to enhance adaptability in dynamic environments, rather than assuming consistent dynamics \cite{zhu2020transfer}.

\subsection{Representation learning}
The quality of state representation plays a crucial role in decision-making performance, as it determines the information content and dynamics of the environment. Effective state representations leverage prior knowledge to improve model performance \cite{battaglia2018relational}. Vehicle state can be represented using continuous variables like position, velocity, and orientation \cite{feng2021intelligent}. The Frenet coordinate system simplifies road structures but may introduce issues such as state confusion and difficulties with varying vehicle numbers. Real-world driving is inherently a Partially Observable Markov Decision Process (POMDP) \cite{bai2015intention}, so the state representation must account for incomplete information and the complexity of policy generation.

Traditional approaches often use images to represent the state space, such as main-view camera images \cite{bojarski2016end} or bird's-eye view images \cite{finn2017deep}, but this can increase the complexity of extracting valid information. Other approaches, like occupancy grids \cite{brechtel2011probabilistic}, process raw data into custom resolution grids, avoiding issues related to dimensional changes. Graph Neural Networks (GNNs) offer an inductive bias model \cite{battaglia2018relational} that uses a flexible graph structure to model complex interactions and enhance decision-making \cite{meirom2021controlling}. GNNs have been successfully applied to solve hidden state inference in POMDPs \cite{ma2021reinforcement}. In this work, a two-layer GNN structure is used to construct a basic decision state, integrating normalized decision-making with local information.

\section{Dynamic Local Enhancement Problem Description}
\subsection{Preliminaries}
Markov decision process (MDP) is commonly used to describe the autonomous driving planning problem. An MDP is defined  by the tuple $\langle \mathcal{S}, \mathcal{A}, \mathcal{R}, \mathcal{T} \rangle$, where $\mathcal{S}$ denotes the state space, $\mathcal{A}$ denotes the action space, $\mathcal{R}: \mathcal{S} \to \mathbb{R}$ is the reward function, and $\mathcal{T}: \mathcal{S} \times \mathcal{A} \to \mathbb{R}$ is the transition probability.  The objective of policy $\pi:\mathcal{S}\times \mathcal{A}\to[0.1]$ is to select actions that maximize the expected reward over time $V_{\pi}(s):=\mathbb{E}_{\pi}[\sum_{i=t}^{H+t} \gamma^{i-t} r_i|s_h=s]$, donated as $V_{\pi}(s)$, where  $\gamma \in [0, 1]$ denotes discount factor, and $H \in \mathbb{N}$ denotes the planning horizon.

In large-scale autonomous driving tasks, treating all environments as a single MDP model may lead to differences between the actual situation and the model from a local region perspective. In this case,  the policy $\pi$ obtained using all data $\mathcal{D}$ of sufficient size $n_{\delta}$   has a performance difference  from the optimal policy $ \tilde{\pi}$ in each real region, the difference donated as:

\begin{equation}
\begin{array}{l}
   \Delta_s V_\pi = ||\Sigma_s V_\pi - \Sigma_s V_{\tilde{\pi}}|| 
\end{array}
\end{equation}

This work focuses on minimizing $\Delta_s V_\pi$ without directly addressing the values of  $n_{\delta}$.

\subsection{Problem description}
The proposed Dynamically Local Enhancement (DLE) planner operates as follows:

A vehicle uses a basic planner for general driving. To adapt to local driving conditions, it collects driving data at its current position and dynamically updates the planner. Once the vehicle leaves the area, the planner reverts to the basic policy. Figure \ref{fig:overview} illustrates the DLE planner’s autonomous driving process.
Our proposed dynamically local enhancement (DLE) planner operates as follows:

\begin{figure}[h]
  \centering
  \includegraphics[width=\linewidth]{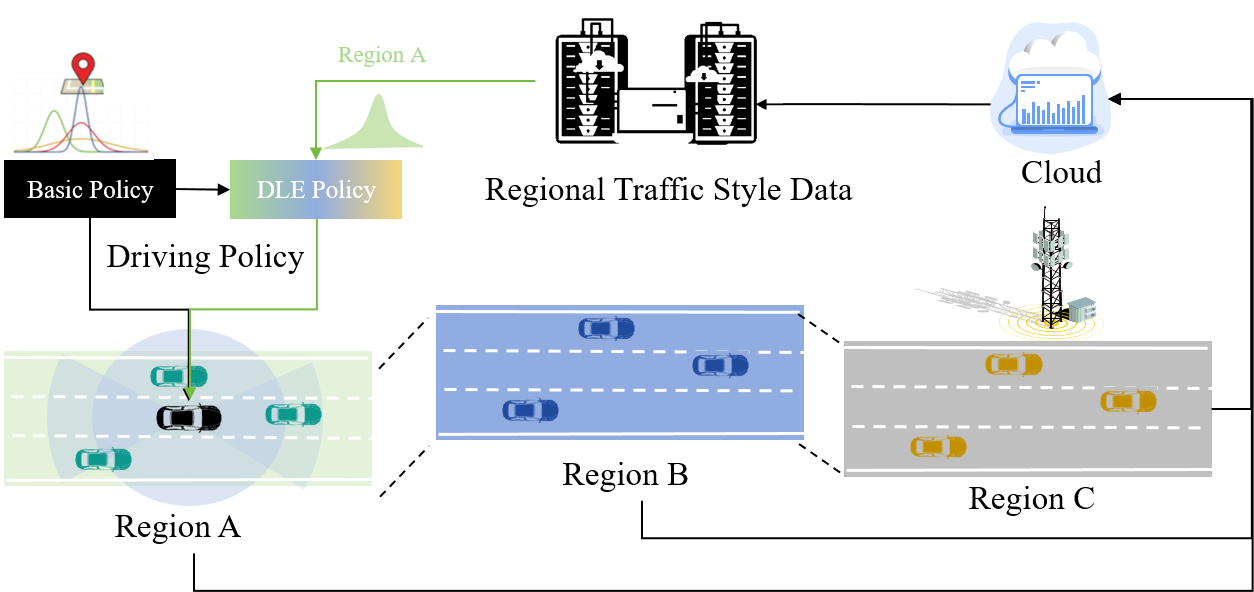}
  \caption{Main idea: Dynamically enhancing a basic driving policy with local driving data when the autonomous vehicle driving to different regions, improving the scalability of the system without significantly expanding the planner’s size.}
  \label{fig:overview}
\end{figure}

This problem can be formalized as:

\begin{equation}
\begin{array}{l}
    \pi(g) = f(\pi_b,\mathcal{D}(g)) \\
    s.t., \Delta_{s(g)} V_{\pi(g)} \leq \Delta_{s(g)} V_\pi
\end{array}
\label{equ: problemdef}
\end{equation}
where $g$ represents the global position parameters.
Equation  \ref{equ: problemdef} shows that the policy only needs to account for future state possibilities at the current position, rather than considering all possible future states. This allows the planner to avoid enlarging the driving model while achieving better performance compared to a general policy.

\subsection{Framework}
\begin{figure}[h]
  \centering
  \includegraphics[width=\linewidth]{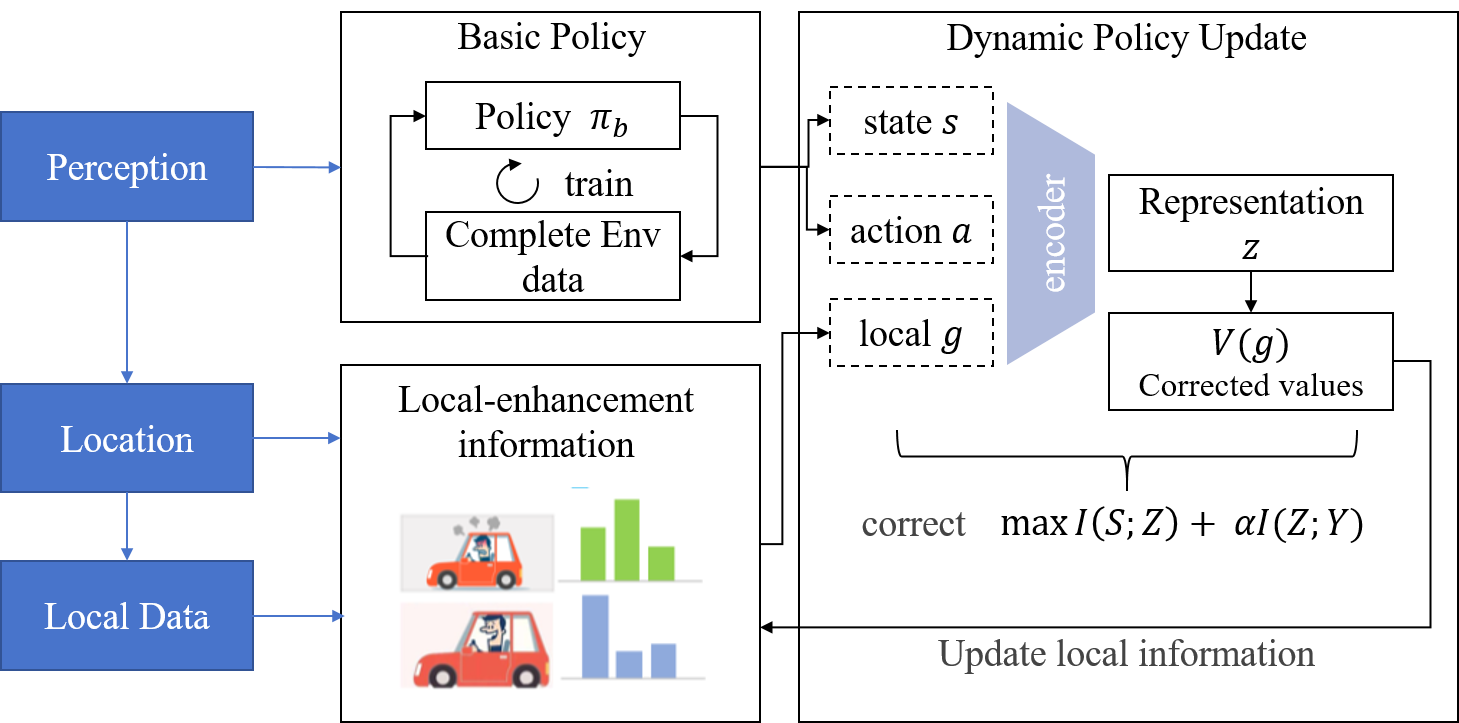}

  \caption{Our framework establishes a hybrid policy architecture that enhance a basic policy with dynamic local adaptation: The basic policy $\pi_b$ works globally, while parallelly, a local-enhancement module extracts region-specific features $g$. Finally, the basic policy will be updated dynamically via information-theoretic optimization. This dual-stream design enables local adaption of driving policy through historical driving data collected locally.}
  \label{fig:framework}
\end{figure}

As shown in Fig. \ref{fig:framework}, our framework consists of three components: the basic policy, the local-enhancement information, and the dynamic policy enhancement. Its goal is to ensure the enhanced policy outperforms the basic one.

The basic policy uses a reinforcement learning model that learns from available data. The local-enhancement module collects regional statistical data, while the dynamic policy enhancement adjusts decisions based on this data and historical performance.

We assume local data can be gathered via road facilities or other vehicles, stored in High-Definition (HD) maps. Several HD maps and driving datasets already support this functionality.

\section{Method}
\subsection{Region-Related Driving Processes}

Traditional MDP models assume that transition probabilities, $\mathcal{T}(s,a,s')$, are independent of geographic location. However, in practice, driving policies that work well in one region may not be suitable for others due to differing driving styles and environmental factors. To address this, we propose the Position-Varying Markov Decision Process (POVMDP), where the transition model is modified to account for the global position $g$: 

\begin{equation}
     \mathcal{T}(s,a,s',g)=Pr(s_{t+1}=s'|s_t=s,a_t=a,G = g)
    \label{equ:transtation2}
\end{equation}

Thus, the POVMDP is defined as $\mathcal{M}(g)=\{\mathcal{S},\mathcal{A},\mathcal{R},\mathcal{T}(g)\}$, and the value function becomes: 
\begin{equation}
\begin{array}{l}
    \qquad V_{\pi}(s,g):=\mathbb{E}_{\pi}[\sum_{t=h}^H \gamma^{t-h}r_t|s_h=s,G=g] 
\end{array}
\label{equ:VQdefine2}
\end{equation}

In this setup, the value function is the weighted average of the POVMDP value functions $V_{\pi}(s,g)$ across regions, which can lead to overestimation or underestimation in different areas.

Directly incorporating the global position into the state space (e.g., $s := \{s, g\}$) increases data sparsity, requiring impractical amounts of data for policy fitting. Furthermore, adapting to new regions through extensive environmental interactions is not feasible.

Our approach mitigates this by using regional data to obtain the value function $V_{\pi}(s,g)$, which helps reduce data sparsity. We transform the problem of directly computing the value function into obtaining an expressive representation space $\mathcal{X}$ that captures the driving environment’s dynamics:
\begin{equation}
\begin{array}{l}
     V(x) = V(u(s,D(g))) = \mathbb{E}_{\pi}[G_t|s_h=s,G=g]
\end{array}
\label{equ:VQdefine3}
\end{equation}
Here, $u$ is a spatial mapping function that maps the state space $\mathcal{S}$ to a new space $\mathcal{X}$ based on data $D(g)$. This approach yields a more compact representation, addressing the challenges of directly incorporating global positions into the state space.

\subsection{Local Enhancement Information Extraction}
To enable adaptive decision-making based on local driving data, we design a basic policy that incorporates regional details, making it suitable for various autonomous driving scenarios. The policy processes decision states, typically represented as images or physical attributes, using a graph network approach \cite{Zhu2021}.

In the basic planner's experience phase, the state \(s\in \mathbb{R}^{n\times F}\) represents the F-dimensional attributes of \(n\) vehicles in the Frenet coordinate system. Each vehicle’s state is expressed in the lane coordinate system (LCS). Environmental features are represented as nodes in the road coordinate system, forming a basic reinforcement learning decision module.

The LCS is constructed with vehicle nodes \(n^v\) and a main reference node \(n^r_0\). Each vehicle node \(n^v_i\) includes attributes such as longitudinal and lateral distances to the centerline (\(s_i, l_i\)), speed \(\dot{s}_i\), lateral speed \(\dot{l}_i\), and orientation \(\Delta \theta_i\). The set of vehicle nodes is denoted as \(N^v=[n^v_0,n^v_1,..,n^v_i]\).

For regional information incorporation, we use graph networks with decision state encoding. Vehicle nodes are encoded and aggregated with other nodes using the same node encoder parameters $\theta_v$:

\begin{equation}
\begin{array}{l}
x^v_i=MLP_{\theta_v}(n^v_i)
\end{array}
\label{equ::xv}
\end{equation}

This ensures normalized coordinate systems. Decision state encoding facilitates reinforcement learning processing of both states with and without regional information. Neural network sparsity maps states to unique representations, with the universal approximation theorem  \cite{hornik1990universal} ensuring efficient mapping. 

Each vehicle node  \(x^{v}_i\) receives related regional information  \(x^r_i\) through the graph network structure. The network parameters  (\(\theta_v\) and \(\theta_r\)) enable an overparameterized mapping. The final encoded decision state $x$ is defined as: \
\begin{equation}
\begin{array}{l}
x=g_\phi(x^v,x^r)=x^v W_0+\underset{j}{\sum}\sigma {x^{r}_{j}}W_1
\end{array}
\label{equ::xg}
\end{equation}

without and with regional information. The implementation is based on the sparse features of neural networks, mapping states with different meanings to unique states. According to the universal approximation theorem \cite{hornik1990universal}, a single mapping function is carried out through the neural network. Each vehicle node \(x^v_i\) receives related local information node \(x^r_i\) through the graph network structure.

The parameters (\(\theta_v\) and \(\theta_r\)) of the network \(MLP_{\theta_v}\) and \(MLP_{\theta_r}\) enable an over-parameterized mapping. The final encoded decision state (\(x\)) is defined as:
\begin{equation}
\begin{array}{l}
x=g_\phi(x^v,x^r)=x^v W_0+\underset{j}{\sum}\sigma {x^{r}_{j}}W_1

\end{array}
\label{equ::xxxx}
\end{equation}

where \(W_0\), \(W_1\) are weight matrices, \(\sigma\) is the nonlinear layer, and \(x^{r}_{j}\) represents related road nodes.

In summary, the basic decision method processes decision states by incorporating vehicle and regional information through graph networks and encoding, ensuring adaptability to diverse traffic environments and supporting any reinforcement learning method. 

\subsection{Dynamically Driving Policy Enhancement}

We use a regional data feedback module to store local traffic characteristics and integrate them into the basic decision algorithm. Local characteristics induce changes in environmental dynamics within a unified state space, with graph neural networks employed to handle interactions between local knowledge features and standardized decision states.

\subsubsection{Regional data Container} 
Autonomous driving maps store geographical data hierarchically, which we leverage as the regional data carrier for local information. This data is encoded using graph networks to represent dynamic differences in the environment. Road nodes are connected based on lane structure and maximum speed distance, forming a network where vehicle nodes interact with the road nodes.

\begin{figure}[h]
  \centering
  \includegraphics[width=0.6\linewidth]{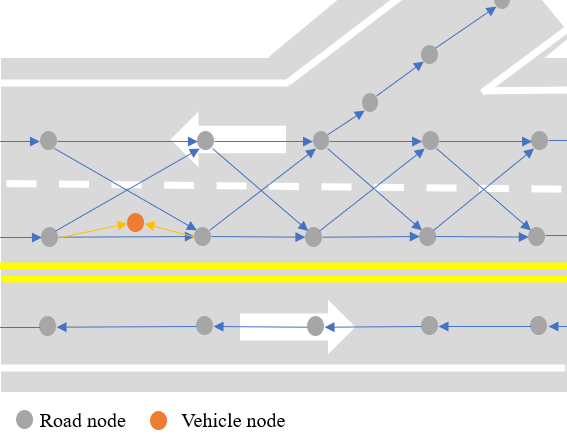}
  \caption{Connection relationships between local feature nodes. Road nodes connect to previous road nodes, and vehicle nodes connect through nodes of interest.}
  \label{fig::nodefeature}
\end{figure}

The connection relationships of road nodes are illustrated in Fig. \ref{fig::nodefeature}. The road node features are defined by their relative position to the reference node $n^r_0$, encoded as:

The connection relationships of road nodes are illustrated in Fig. \ref{fig::nodefeature}. Lane nodes align along the road's center line, with adjacent points in the same lane set at half of the 1s travel distance at the maximum speed limit. 
Each node connects to its predecessor node, and road nodes connect to their associated vehicle nodes.
\begin{equation}
\begin{array}{l}
h^r_i=MLP_{\theta r}(n^r_i-n^r_0)
\end{array}
\label{equ::hr}
\end{equation}

Graphsage is used to aggregate information from neighboring nodes \cite{Ahmed2017}. The aggregation process is: 

\begin{equation}
\begin{array}{l}
    x^k_{N_{(n)}}=\sigma (MEAN(x^{k-1}_u,\forall u\in N(n))\\
    x^k_v=\sigma(W^k\cdot CONCAT(x_v^{k-1},x^k_{N_{(n)}}))
    \label{equ:X_v'}
\end{array}
\end{equation}

where \(MEAN\),  \(CONCAT\) denote the mean clustering function and the vector concatenation function, respectively.
\(N_{(n)}\) denotes the neighborhood node connected to node \(n\). 
\(\sigma\), \(W^k\) and \(k\) denote the non-linear function, the weight matrix, and the depth of collection. The state \(x^k\) at depth \(k\) is defined as \(x^r\), forming local characteristic parameters \(x\) with Eq. \ref{equ:X_v'}.

\subsubsection{Historical Information Feedback}

\begin{figure}[h]
  \centering
  \includegraphics[width=\linewidth]{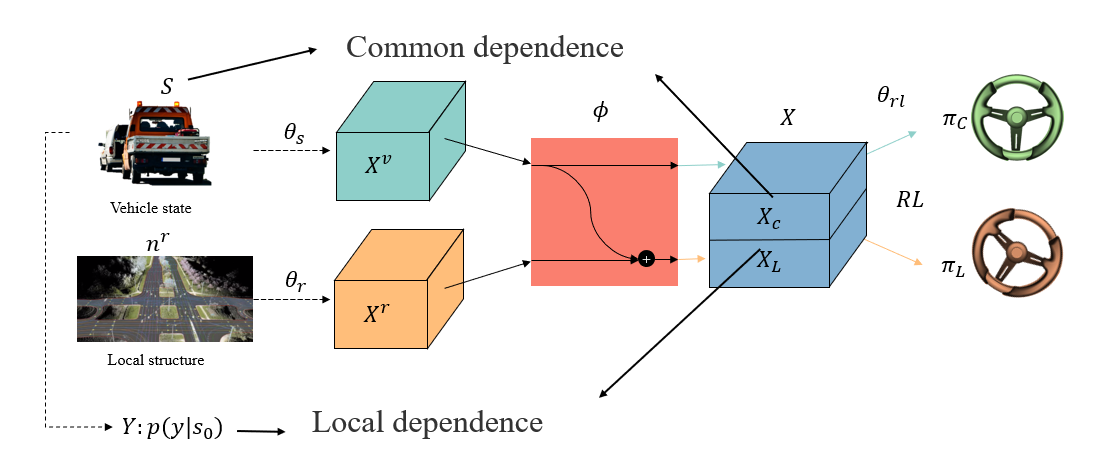}
  \caption{Local information extraction process.}
  \label{fig:porcess}
\end{figure}

As shown in Fig.\ref{fig:porcess}, the policy state $x$ is divided into $x_c$ and $x_l$, where $x_c=g_\phi(x^v)$ is the state without local info , and $x_l=g_\phi(x^v,x^r)$ indicates the combined information. 
The mutual information between the state-action trajectory $Y: y=[s_0,a_0,s_1, ..., s_h]$ and local information $X_c=g_\phi(E_{\theta_s}(s))$  is maximized to reduce uncertainty in decision-making. 

Local mutual information $I(Y; X_l)$ is estimated using theMutual Information Neural Estimation \cite{belghazi2018mutual} method,  and common mutual information  $I(S; X_c)$ is estimated similarly.

\begin{equation}
\begin{array}{l}
I(Y; X_l) := D_{KL}(P_{Y, X_l} || P_Y P_{X_l}) \geq \widehat{I}_{w}^{\left( DV \right)}\left( Y; X_l \right) \\
:= \mathbb{E}_{P_{Y, X_l}}\left[ T_w(y, x_l) \right] - \log \mathbb{E}_{P_Y P_{X_l}}\left[ e^{T_w(y, x_l)} \right]
\end{array}
\end{equation}
where $T_w: Y \times X_l \to \mathbb{R}$ is a function modeled by parameter $w$. $X_l=g_\phi(E_{\theta_v}(s),E_{\theta_r}(n^r))$. Local mutual information $I(Y; X_l)$ is estimated by maximizing $\widehat{I}_{w}\left( Y; X_c \right)$:

\begin{equation}
\begin{array}{l}
w_l, \varphi, \theta_s, \theta_r = \underset{w_l, \varphi, \theta_s, \theta_r}{\arg \max} \widehat{I}_{w_{l}}\left( Y; g_\phi(E_{\theta_v}(s), E_{\theta_r}(n^r)) \right)
\end{array}
\end{equation}
where $w_l$ denotes local function parameters. $\varphi, \theta_s, \theta_r$ are the parameters of the decision state encoder, the vehicle, and the local structure encoder mentioned earlier.

Common mutual information $I(S; X_c)$ is estimated in the same way with common function parameter $w_c$:

\begin{equation}
\begin{array}{l}
w_c, \varphi, \theta_s = \underset{w_c, \varphi, \theta_s}{\arg \max} \widehat{I}_{w_c}\left( S; g_\phi(E_{\theta_s}(s)) \right)
\end{array}
\end{equation}

The two objectives update the three state encoding networks before the input to the decision system, so they are updated together:

\begin{equation}
\begin{array}{l}
\underset{w_l, w_c, \varphi, \theta_s, \theta_r}{\arg \max} (\widehat{I}_{w_c}\left( S; X_c \right) + \alpha \widehat{I}_{w_{l}}\left( Y; X_l \right) )
\end{array}
\end{equation}
where $\alpha$ is a hyperparameter. The difference from the original method is that the data is constantly updated and changed after the planning process.

\subsubsection{Training Process}

\begin{algorithm}
\label{AEAM}
\caption{DLE Planner}
\begin{algorithmic}[1]
\State Initialize replay memory $D$, $\widetilde {\mathcal{D}}$
\State Initialize action-value function $Q_{\theta_{rl}}$, $w_l, w_c, \varphi, \theta_s, \theta_r$
\While{episode $<$ $M$}
    \State Initial state $s_0$
    \State Choose to use local information with a probability of 0.5
    \While{not Done}
        \State $
        x=\begin{cases}
        	g_\phi(E_{\theta_s}(s)), \text{without local info}\\
        	g_\phi(E_{\theta_s}(s), E_{\theta_r}(n^r)), \text{with local info}\\
        \end{cases}$
        \State Select action $a_t$ with $\varepsilon$-greedy with input $x$
        \State Observe and store $(x_t, a_t, r_t, x_{t+1})$ into $\mathcal{D}$
        \State Encoding and store $(y, x_l, x_c, s)_t$ into $\widetilde {\mathcal{D}}$
        \State Draw minibatch samples from joint distribution $P_{s, x_c}$, $P_{y, x_l}$ and marginal distribution $P_s$, $P_{x_l}$
        \State Update $w_l, w_c, \varphi, \theta_s, \theta_r$ with target Eq.\ref{equ::L_enb}
        \State Sample minibatch from $D$
        \State Update $\theta_{rl}$, $w_l, w_c, \varphi, \theta_s, \theta_r$ with target Eq.\ref{equ::L_rl}
    \EndWhile 
\EndWhile 
\State \textbf{End While}
\end{algorithmic}
\end{algorithm}

We take the DQN algorithm as an example to illustrate the update method. With two replay buffers: one for storing the reinforcement learning transitions and one for storing the corresponding encoded states. The loss function for reinforcement learning is: 
\begin{equation}
\begin{array}{l}
L_{rl}(\theta_{rl}, \theta_{enb})=\mathbb{E}\left[ (r+\gamma \underset{a‘}{\max}Q\left( s', a' \right) -Q\left( s, a \right) ^2 \right] 
\end{array}
\label{equ::L_rl}
\end{equation}
where $\theta_{rl}$ denotes the parameters of Q-net and $\theta_{enb}$ denotes all encoding parameters. The loss of the decision also updates the encoding network, while the encoding network performs asynchronous updates. Changes in the planned outcomes will affect the dynamics of the environment, resulting in fluctuations in planning. To eliminate the oscillation caused by such updates, the loss function of the encoding part of the network is modified as:

\begin{equation}
\begin{array}{l}
L_{enb}(w_l, w_c, \varphi, \theta_s, \theta_r) = - \beta(\widehat{I}_{w_c}\left( S; X_c \right) + \alpha \widehat{I}_{w_{l}}\left( Y; X_l \right))
\end{array}
\label{equ::L_enb}
\end{equation}
where $\beta$ is a hyperparameter that drops to 0 over the training process , allowing the system to stabilize in later stages.

\section{Experiment \label{sec:eva}}

This section aims to conduct a comprehensive evaluation of the proposed dynamically local enhancement planner (DLE) by comparing it with planners that do not take local information into account.
We will first introduce the region-related test scenarios, the design of baseline planners, and the performance metrics. Subsequently, the training process will be presented. Finally, we will assess the performance of the proposed planner.

\subsection{Design of Region-related Test Scenarios}

\begin{figure}[h]
  \centering
  \includegraphics[width=\linewidth]{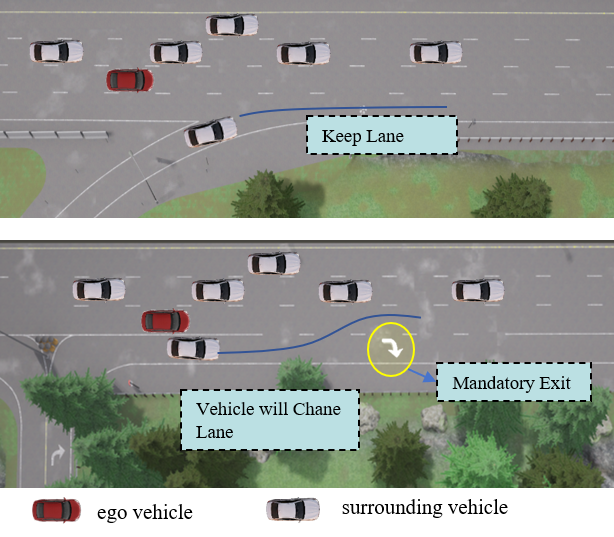}
  \caption{The test scenarios sourced from different regions are designed to have distinct road structures and surrounding vehicle behaviors. In the upper scenario, the merging - in vehicle is likely to maintain its lane. Conversely, in the lower scenario, it is more prone to changing lanes.}
  \label{fig:scenario}
\end{figure}

To validate the adaptability of the proposed method to region-specific driving scenarios, two test scenarios are designed from different regions, as shown in Fig. \ref{fig:scenario}. In these scenarios, the autonomous vehicle interacts with merging vehicles, where road structures and vehicle behaviors differ between regions. Specifically, the lane-changing probabilities of merging vehicles vary; in one region, they are more likely to change lanes in front of the autonomous vehicle, creating different optimal driving policies for each scenario. This tests whether the trained policy can adapt to the unique characteristics of each region. Additionally, within each scenario, the initial positions of vehicles and their behaviors are generated using probabilistic sampling, introducing further variability.

The scenarios are implemented in the highway-env simulator based on real traffic data \cite{highway-env}. The longitudinal behavior of the environmental vehicle is modeled using the Intelligent Driver Model (IDM) \cite{treiber2000congested}, and the lateral behavior is modeled using the Minimizing Overall Braking Induced by Lane Change Model (MOBIL) \cite{Paden2016}. The  parameters of the IDM and MOBIL model are shown in Table.\ref{tab::idm}.

\begin{table}[h]  

\centering  
   \caption{Parameters of IDM and MOBIL model}  
\setlength{\tabcolsep}{5mm}{
 \begin{tabular}{p{3cm}cc} 
     \toprule
       Parameters & Symbol & Value \\ [0.9ex]  
        \hline
        Desired velocity      & ${\dot x}_0$    & 50 $km/h$ \\
        Desired time gap      & $T$             & 1.5 $s$\\
        Jam distance          & $g_0$           & 5 $m$\\
        Exponent for velocity & $\delta$        & 3.4-4.5 \\
        Max acceleration      & $a$             & 3 $m/s^2$\\
        Desired deceleration  & $b$             & -5 $m/s^2$\\
        Politeness index      & $p$             & 0.2\\
        Lane change threshold & $\delta a_{th}$ & 0.2 $m/s^2$\\
         \toprule
   \end{tabular}
}
   \label{tab::idm}
\end{table}

We use the positions, speeds, and orientations of the seven nearby vehicles in the lane coordinate system as the state space, with discrete actions consisting of acceleration, deceleration, lane change, and hold.
Both testing scenarios shares the same set of reward function, which consists of four parts: collision penalty $r_c$, velocity reward $r_v$, lane change penalty $r_l$, and politeness $r_p$, as follows:

\begin{equation}
    \begin{array}{l}
       r= w_c\cdot r_c + w_v\cdot r_v + w_l\cdot r_l + w_p\cdot r_p
    \end{array}
    \label{equ:reward}
\end{equation}
where $w$ is the weight. When a collision occurs, $r_c = 1$; otherwise, it is $0$. $r_v$ is normalized to $[0,1]$ according to the maximum speed limit of the current road. $r_l = 1$ when the lane change action is successfully executed. $r_l$ is used to reduce the impact of the self-vehicle on the environmental vehicle and is defined as the proportion of the environmental vehicle speed reduction caused by the ego-vehicle.
\begin{equation}
    \begin{array}{l}
       r_v= \frac{(v_{target}-v)}{v_{target}}
    \end{array}
    \label{equ:vl}
\end{equation}
where $v_{target}$ is the target speed of the surrounding vehicle. The weight of each part is $w_c = -1$, $w_v = 0.2$, $w_l = -0.05$, and $w_p = -0.1$,with a discount factor  $\gamma=0.95$.

\subsection{Baseline planner design}
We compare the proposed DLE model with two baselines: a single-model (SM) driving policy of the same parameter size, and a large-parameter global model.

\subsubsection{Large-Parameter Global Driving Policy}
The first baseline verifies if a large-parameter model can adapt to diverse regional driving characteristics. Drawing on the Mixture of Experts (MoE)\cite{shazeer2017outrageously} in large language models, we designed a large parameter global driving policy (GM). It selectively activates a specific expert policy according to location, combining regional expert policies into a large-parameter driving model. In this way, the parameter number of $GM$ policy is $n_s$(number of regions) times of the proposed DLE planner.

\subsubsection{Local Driving Policy}
We employ the deep Q-learning (DQN) method to generate local driving policies. For comparison, we design three local policies: $LM_1$, trained solely in region 1; $LM_2$, trained in region 2; and $LM_{12}$, trained using data from both regions. The parameter count of the single-model policies is consistent across these cases. The training data for each baseline are shown in Table \ref{tab::data2}.

\begin{table}[h]  

\centering  
   \caption{Comparsion of Different Baselines}  
\setlength{\tabcolsep}{2mm}{
 \begin{tabular}{p{1.8cm}p{2cm}p{1cm}p{1cm}p{1cm}} 
     \toprule

       Model &      parameters \qquad number& Data  1 & Data 2  & Global \qquad Location \\ [0.9ex]  
        \hline

        $LM_1$           &$9.6\times 10^6$& \checkmark     & $\times$           &$\times$   \\
        $LM_2$           &$9.6\times 10^6$& $\times$        & \checkmark           &$\times$   \\
        $LM_{12}$        &$9.6\times 10^6$& \checkmark     & \checkmark           & $\times$  \\
        $GM$   &$n_s\times9.6\times 10^6$& \checkmark     & \checkmark        &$\checkmark$   \\
        \textbf{DLE}     &$9.6\times 10^6$& \checkmark     & \checkmark        &\checkmark    \\
        \toprule

     \end{tabular}
}
\label{tab::data2}
\end{table}

\subsection{Performance metrics}
\subsubsection{Average performance ratio (APR)}

To evaluate policy performance across regions, we define APR as:

\begin{equation}
    APR=\frac{\sum^{n_s}_{i=1}{G^i_{\pi}}}{\sum^{n_s}_{i=1} G^i_{opt}}
    \label{metric}
\end{equation}

where $n_s$  is the number of regions, $G^i_{\pi}$ is the average reward of the target policy in the local region $i$, $G^i_{opt}$ is the average reward of optimal policy in the local region $i$.
Optimal policy refers to the policy that interacts only with the local interaction environment.

\subsubsection{Collision Rate}
The collision rate $R_c$ measures the frequency of collisions of a driving policy during the evaluation testing, which is defined as:

\begin{equation} 
R_c = \frac{N_C}{N_T} 
\end{equation}

\subsection{Result}

\begin{figure}[h]
  \centering
  \includegraphics[width=1\linewidth]{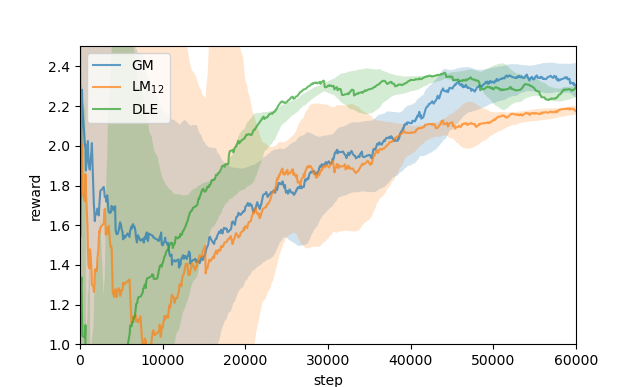}
  \caption{Training Process Episode Reward. The solid line represents the mean episode reward across the two scenarios, while the shaded area indicates the 90\% confidence interval.}
  \label{fig:train}
\end{figure}

The training process is shown in Figure \ref{fig:train}. To evaluate and compare the overall performance across regions, we present the average performance of $LM_{12}$, $GM$, and $DLE$ Planner in two scenarios. The solid line represents the mean episode reward from 10 random tests during training, with the shaded area indicating the 90\% confidence interval.

During training, the $GM$ method converges more slowly due to its large parameter size and sparse updates. The $LM_{12}$ model, trained with mixed data from both regions, lags behind the other methods in final performance.

\begin{table}[h]  

\centering  
   \caption{The average performance ratio results of different methods}  
   \setlength{\tabcolsep}{5mm}{
 \begin{tabular}{p{0.8cm}p{1cm}p{1cm}p{0.1cm}p{1cm}} 
     \toprule

     Driving Policy & Training Scenario & Test Scenario & $APR$ & Collision Rate $R_c$  \\ [0.9ex]  
    \hline
   $GM$ &  1\&2 & 1\&2& 1.00 &0\% \\
    \multirow{2}{*}{$LM_1$}& 1& 1\&2& 0.70 & 36\% \\
    ~& 1& 2 & 0.41 & 72\% \\
    \multirow{2}{*}{$LM_2$} & 2 & 1\&2& 0.90 & 0\% \\
    ~&2& 1& 0.81 & 0\% \\
    $LM_{12}$ & 1\&2&1\&2& 0.91 & 6\% \\

        \textbf{DLE} &1\&2&1\&2& \textbf{0.99} & \textbf{0\%} \\

        \toprule
   \end{tabular}
}
\label{tab1}
\end{table}
We conducted 100 tests for each policy in both scenarios, summarized in Table \ref{tab1}. The $GM$ model's APR value was used as the benchmark (1.0) for comparison.

The $LM_1$ model, trained on Scenario 1 alone, showed higher collision rates in Region 2, causing a drop in APR. The $LM_2$ model avoided collisions but saw performance drop to 0.90. The $LM_{12}$ model, trained on both regions, performed slightly better than the first two but still fell short of $GM$. The trade-off in performance across regions highlights that simply increasing training data diversity does not always improve planning performance.

In contrast, the DLE model outperformed the single-region models and closely approached $GM$'s performance. The DLE planner proves superior in dynamic environments by leveraging storage space to optimize planning results, surpassing general planners.

\begin{figure}
\begin{minipage}[b]{1\linewidth}
    \centering
    \subfloat[][Example Case of the LM planner]{
    \label{fig:result1}
    \includegraphics[width=\linewidth]{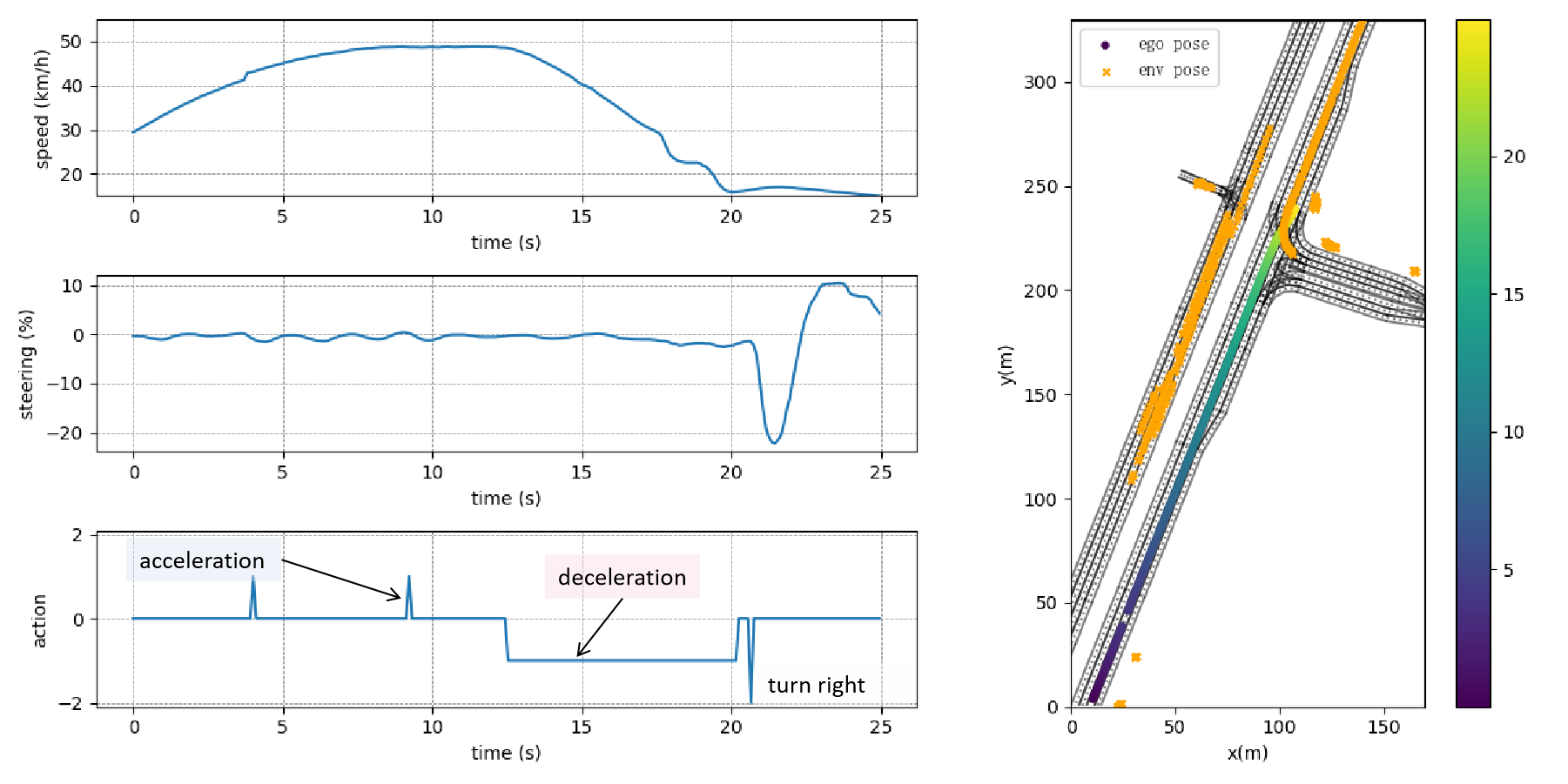}}
\end{minipage}
\medskip

\begin{minipage}[b]{1\linewidth}
    \centering
    \subfloat[][Example Case of the proposed DLE planner]{
    \label{fig:result2}
    \includegraphics[width=\linewidth]{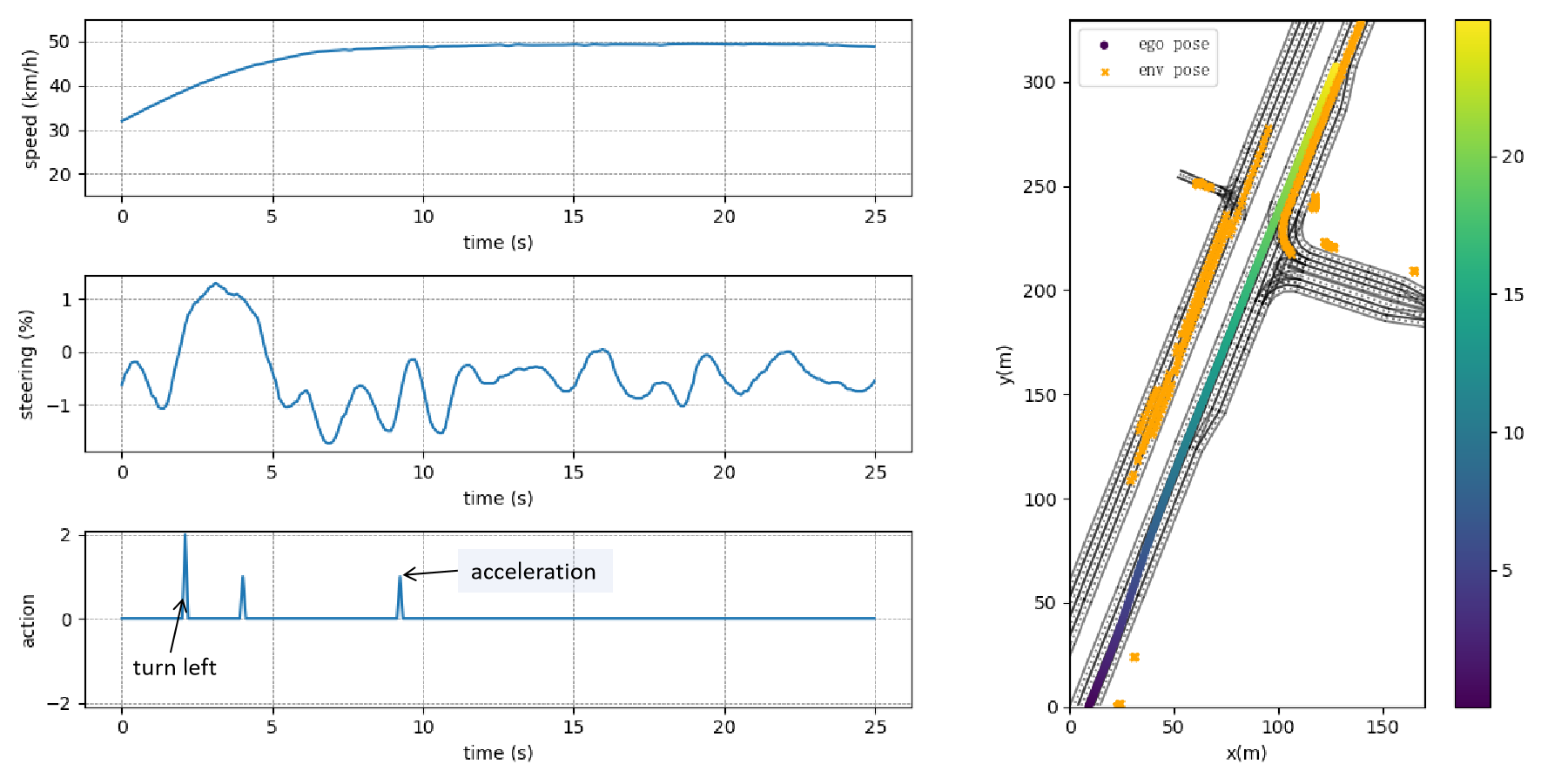}}
\end{minipage} 

\caption{Test results in a Specific Region.The upper subgraph illustrates the planning outcomes of $LM$ planner, while the lower subgraph displays the planning outcomes with the inclusion of extracted local driving feature. Distinct decision actions emerge within the state space described by the lane coordinate system.}

\label{fig:result}
\end{figure}

We show a case to compare the proposed DLE planner with a trained $LM_{12}$ driving policy at a T-junction.The autonomous vehicle drives straight on the main road, while an environmental vehicle may turn left from the intersection. Both policies gather information about surrounding vehicles, and the DLE planner uses historical driving data at the intersection.

Figure \ref{fig:result} shows the ego vehicle's trajectory, speed, steering angle, and behavioral decisions. The $LM_{12}$ planner initiates deceleration upon encountering merging vehicles and switches lanes to the left to avoid them. However, half of the vehicles in the right lane do not interact with the ego lane, causing the policy to prioritize deceleration.

The DLE planner, in contrast, maintains maximum speed and preemptively changes to the left lane to avoid potential encounters with merging traffic. This approach minimizes lane-change losses and reduces stop-and-go occurrences, showcasing the DLE planner's ability to adapt to local traffic dynamics by incorporating historical behavior into its decisions.

\section{Conclusion}

In this study, we proposed the DLE planner, a model designed to enhance the adaptability of autonomous driving policies across regions with diverse driving characteristics. By leveraging a regional driving model that integrates local environmental data and historical driving behavior, the DLE planner demonstrates superior performance compared to traditional approaches, such as single-region models and large-parameter global models. 
The DLE planner strikes a balance between generalization and localization, closely matching the performance of the large-parameter model while maintaining more efficient training and decision-making processes. The ability to incorporate historical data allows the DLE planner to optimize decisions in dynamic environments, reducing collision risks.

Overall, the DLE planner proves to be a robust and scalable solution for autonomous driving, offering enhanced flexibility and performance across diverse driving scenarios, and has the potential to enable large-scale deployment of autonomous vehicles without the need for substantial expansion of on-device driving models.



\ifCLASSOPTIONcaptionsoff
  \newpage
\fi

\bibliographystyle{IEEEtran}
\bibliography{main}

\ifCLASSOPTIONcaptionsoff
  \newpage
\fi

\end{document}